\documentclass{article}
\usepackage[utf8]{inputenc}  
\usepackage[T1]{fontenc}     
\usepackage{amsmath}         
\usepackage{ijcai25}

\usepackage{times} 
\usepackage{soul}
\usepackage{url}
\usepackage[hidelinks]{hyperref}
\usepackage[utf8]{inputenc}
\usepackage[small]{caption}
\usepackage{graphicx}
\usepackage{amsmath}
\usepackage{amsthm}
\usepackage{booktabs}
\usepackage{algorithm}
\usepackage{algorithmic}
\usepackage[switch]{lineno}
\usepackage{multirow}
\usepackage[table]{xcolor}
\usepackage{epstopdf}

\definecolor{customcolor}{RGB}{254,247,242}
\definecolor{green}{RGB}{ 0,136,51}
\definecolor{blue}{RGB}{0,102,204}
\definecolor{red}{RGB}{ 202,0,0}

\newcommand{\taskname}{Class-Guided Camouflaged Object Detection}
\newcommand{\methodname}{CGNet}
\newcommand{\datasetname}{CamoClass}
\newcommand{\CPGname}{class prompt generator}
\newcommand{\Detetorname}{class-guided detector}

\newenvironment{authorfoot}
  {}
  {}

\title{CGCOD: Class-Guided Camouflaged Object Detection}

\author{
Chenxi Zhang$^1$\and
Qing Zhang$^1$\protect\footnotemark[1]\and
Jiayun Wu$^1$ \and
Youwei Pang$^2$\protect\footnotemark[1]\\
\affiliations
$^1$ Shanghai Institute of Technology,
$^2$ Dalian University of Technology \\
\emails{\normalsize
\texttt{\{236142153\}@mail.sit.edu.cn},
\texttt{lartpang@gmail.com}
}
}

\begin{document}

\maketitle
\begin{authorfoot}
  \footnotetext[1]{Corresponding author.}
\end{authorfoot}

\begin{abstract}
Camouflaged Object Detection (COD) aims to identify objects that blend seamlessly into their surroundings. 
The inherent visual complexity of camouflaged objects, including their low contrast with the background, diverse textures, and subtle appearance variations, often obscures semantic cues, making accurate segmentation highly challenging. 
Existing methods primarily rely on visual features, which are insufficient to handle the variability and intricacy of camouflaged objects, leading to unstable object perception and ambiguous segmentation results. 
To tackle these limitations, we introduce a novel task, class-guided camouflaged object detection (CGCOD), which extends traditional COD task by incorporating object-specific class knowledge to enhance detection robustness and accuracy. 
To facilitate this task, we present a new dataset, \datasetname, comprising real-world camouflaged objects with class annotations. 
Furthermore, we propose a multi-stage framework, \methodname, which incorporates a plug-and-play \CPGname~and a simple yet effective \Detetorname. 
This establishes a new paradigm for COD, bridging the gap between contextual understanding and class-guided detection. 
Extensive experimental results demonstrate the effectiveness of our flexible framework in improving the performance of proposed and existing detectors by leveraging class-level textual information. Code: https://github.com/bbdjj/CGCOD.

\end{abstract}

\section{Introduction}
\label{sec:introduction}

Accurate object perception in visual scenes is essential for image-level semantic reasoning and cognition.
However, camouflaged objects often blend seamlessly into their surroundings, obscuring their appearance or shape, which poses a significant challenge for object perception.
This makes capturing the semantic features of camouflaged objects particularly challenging.
Addressing this issue holds significant value in domains such as
medical image segmentation \cite{fan2020pranet},
smart agriculture \cite{rustia2020application},
and environmental protection \cite{nafus2015hiding}.

\begin{figure}[!t]
\centering  
\includegraphics[width=\linewidth]{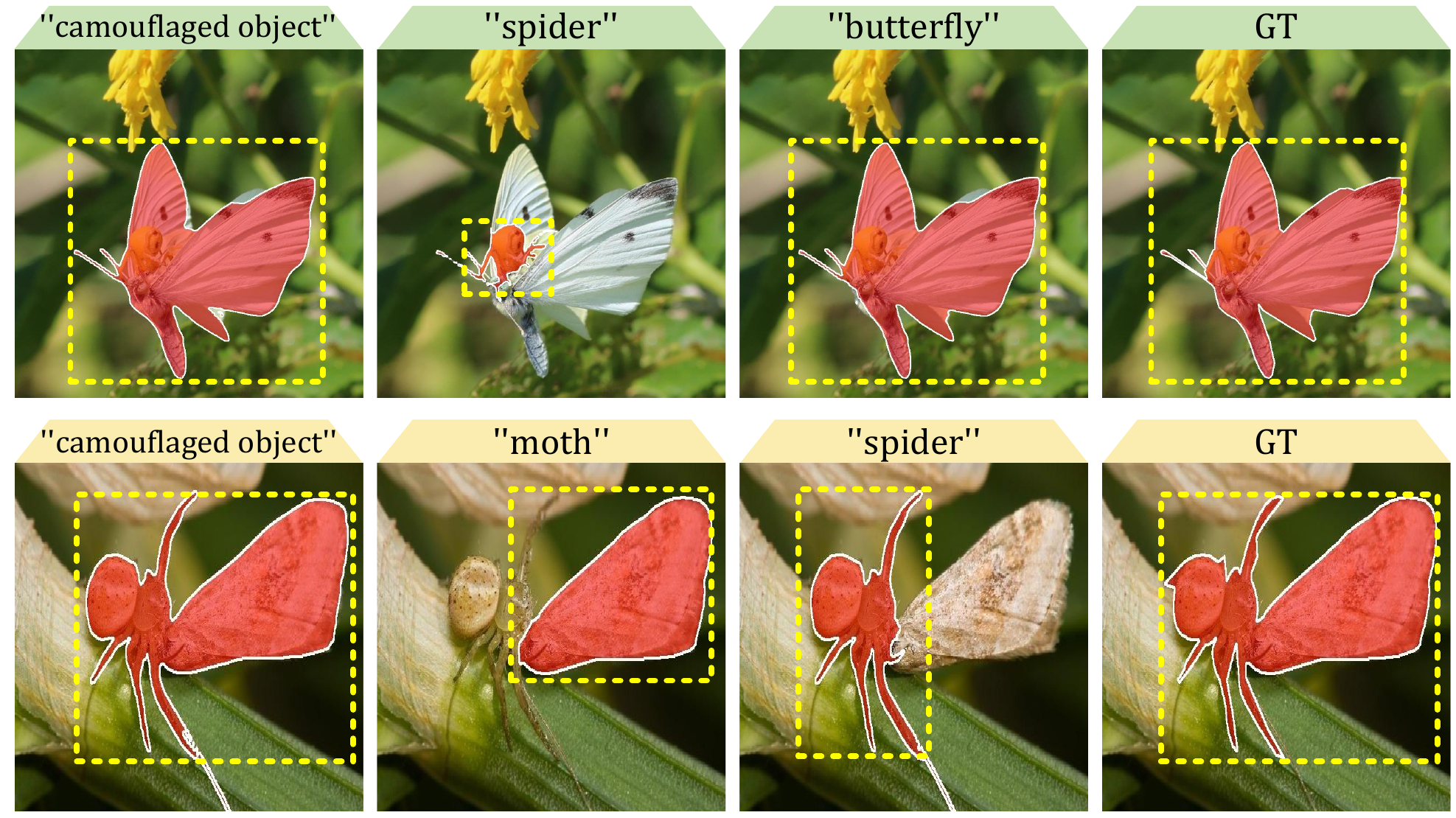}
\caption{
Visualization of the proposed \methodname\ illustrating its segmentation performance guided by various textual information.
}
\vspace{-13pt}
\label{fig:directional}
\end{figure}

In the early stages of COD, algorithms primarily rely on hand-crafted features to perform the detection process \cite{galun2003texture}, which exhibit limited generalization capabilities.
With the rise of deep learning, existing COD methods have significant advanced by incorporating prior information such as
edge \cite{BGNet},
gradient \cite{zhang2023collaborative},
and uncertainty \cite{UEGD},
or employing multi-task learning \cite{FEDER2023} to capture rich semantics.
Despite substantial advancements across various scenes, these methods still face significant challenges, including limited robustness, low detection accuracy, and inadequate generalization capability.  We argue that these challenges can be attribute to two aspects: (1) the inherently inconspicuous nature of camouflaged objects, which allows them to blend seamlessly into their surroundings, and (2) the complexity and variability of the environment conditions, such as occlusion and background saliency, which further hinder accurate detection. 


These challenges arise from relying solely on purely visual paradigm, which fail to capture the full semantic characteristics of camouflaged objects. This limitation becomes particularly pronounced in scenes where objects closely resemble their backgrounds or are heavily occluded. 
Camouflaged objects exhibit strong visual similarity to their background (e.g., shape, color, scale), yet maintain significant semantic differences.
However, existing benchmark datasets primarily focus on visual information and lack stable, explicit information to effectively emphasize these semantic differences. As a result, purely visual models often suffer from high rates of false positives (e.g., misclassifying background regions as objects) and false negatives (e.g., failing to detect actual objects), leading to a substantial decline in detection accuracy.
Moreover, various factors such as environmental lighting, interference, and occlusion, can profoundly affect the visual appearance of the object, making it challenging for models to consistently extract distinctive object features.
For instance, partial occlusion distorts the original structure and contour of the object, reducing the model’s ability to recover complete semantic information from limited visual cues.

In recent years, integrating visual and textual features has proven effective for enhancing image content understanding, thereby improving performance in various multi-modal tasks \cite{tan2019lxmert}.
Targeted textual prompts provide explicit semantic cues that help models focus on object-related features and reduce background interference \cite{radford2021learning}. 
However, existing COD benchmark datasets provide only visual information, while current visual-textual COD methods \cite{ACUMEN} depend on coarse-grained text generated by large language models.
Therefore, it is of great significance for the COD community to build a dataset that integrates both visual and textual information and develop a new COD framework.

Motivated by the above discussion, we introduce a new task, \textbf{\taskname\ (CGCOD)}.
This task accurately focus on relevant regions that likely contain camouflaged objects by leveraging the specific class text, thereby improving model's detection accuracy and robustness.
To support our research, we construct a new dataset, \textbf{\datasetname}.
It integrates multiple COD benchmark datasets, including CAMO \cite{CAMO}, COD10K \cite{SINetV2}, CHAMELEON \cite{CHAMELEON}, and NC4K \cite{NC4K}.  
We annotate all images with their corresponding class information.  
To further explore the role of class guidance in COD, we propose a new method, \methodname, which consist a plug-and-play {\CPGname} (CPG) and a novel high-performance baseline {\Detetorname} (CGD).
The CGD is designed to seamlessly integrate with CPG. 
And we introduce a class semantic guidance  as a bridge between CPG and CGD to fuse information from multiple sources and reduce domain differences between them.
Fig.~\textcolor{red}{\ref{fig:directional}} shows the ability of {\methodname} for the CGCOD task.

%
In the CPG, we introduce a cross-modal multi-head attention module to enhance class semantics by facilitating interactions between visual and textual features. 
Additionally, our multi-level visual collaboration module (MVCM) ensures comprehensive alignment and integration of these features at various levels, enabling cooperative extraction of object semantics.
Together, these modules generate class prompt feature $G_c$ that guide the COD detector for precise segmentation. 
In the CGD, we design a semantic consistency module (SCM) to reduce false negatives and positives by leveraging the class prompt features from the CPG.
CGNet further explores multi-modal feature correlations, suppresses background noise, and achieves accurate predictions.

The main contributions of our paper are as follows:
\begin{itemize}
\item We propose a new task, \textbf{CGCOD}, which is the first time to fully explore the performance of COD methods under the class guidance and provide a new insight for the traditional COD field.
\item We construct a novel dataset \textbf{\datasetname}~by integrating existing COD datasets and annotate the class label of the camouflaged object.
\item We design \methodname, a new method featuring a plug-and-play \CPGname~(CPG) that enhances existing COD detectors, and a high-performance baseline \Detetorname~optimized for the CPG.
\item Extensive experiments demonstrate the superior performance of our \methodname over existing state-of-the-art methods. Further, the CPG's contribution to existing detectors' performance demonstrates its strong pluggability.
\end{itemize}

\section{Related Works}
\label{sec:related_works}

\noindent\textbf{Referring Image Segmentation (RIS).}
RIS is a fundamental computer vision task aiming to segment object regions in an image based on natural language descriptions.  
Early RIS methods are often built upon CNN-LSTM architectures~\cite{hu2016segmentation}.
They encode image and text information separately, and leverage simple attention mechanisms.  
This limits cross-modal interaction, making it challenging for models to selectively focus on object-relevant features, resulting in coarse segmentation.  
For example, \cite{CRIS} introduces a vision-language decoder that maps fine-grained text semantics to pixel-level representations for improved modality alignment. \cite{CGFormer} employs a mask classification framework with token-based querying to enhance object-level understanding and cross-modal reasoning. \cite{ASDA} uses a word-guided dual-branch structure to dynamically align visual features with linguistic cues, boosting RIS performance. 
%
Although our method also integrates textual and visual modalities, our objective differs substantially from traditional RIS.
RIS tasks use natural language to segment clearly described objects, whereas our task combines object class knowledge to detect camoflaged objects blending into the background, making it more challenging. 

\noindent\textbf{Camouflaged Object Detection (COD).}
In nature, animals use camouflage as a defense mechanism to evade predators~\cite{camouflaged_mechanism}.
The high complexity of camouflaged strategies makes COD a consistently challenging task.
Early COD methods relied on handcrafted features like texture~\cite{galun2003texture}, gradient~\cite{zhang2016bayesian}, and color~\cite{color} to distinguish foreground from background but suffered from limited feature extraction capabilities, especially in complex scenes.
In contrast, deep learning-based approaches, such as CNNs and Transformers, automatically learn camouflaged object features, offering significant improvements.
For example, \cite{SINet} uses densely connected features and receptive fields to mimic a predator's search and recognition process. \cite{zoomnet2022} integrates multi-scale features and hierarchical units to emulate human visual patterns when observing blurry images, capturing hybrid-scale semantics. 
\cite{BGNet} leverages edge cues to guide the model in perceiving object structures.
\cite{FSNet} introduces a non-local token enhancement mechanism and a feature compression decoder to enhance feature interactions. \cite{FEDER2023} employs a learnable wavelet transform to decompose features into different frequency domains, effectively distinguishing foreground from background.
With the emergence of large models, \cite{DSAM} utilizes SAM's zero-shot capability by combining depth and RGB features for precise segmentation. Additionally, \cite{ACUMEN,ruan2024mm} use environment descriptions generated by large language models to investigate how background camouflage attributes affect object detection. These approaches demonstrate deep learning's flexibility and innovation in the COD task, offering diverse solutions.
Despite many advances, existing COD methods rely too heavily on visual features, which fail to adequately address the complex characteristics of camouflaged objects.
This limitation may not only lead to poor model robustness, but also cause misrecognition problems.
In contrast, our work introduces class-level textual information into the COD domain, along with a new benchmark dataset.
By explicitly incorporating class semantics, our model can more accurately focus on the object region in the image and effectively reduce the interference of the complex environment on object perception.


\begin{figure}[!t]
\centering  
\includegraphics[width=\linewidth]{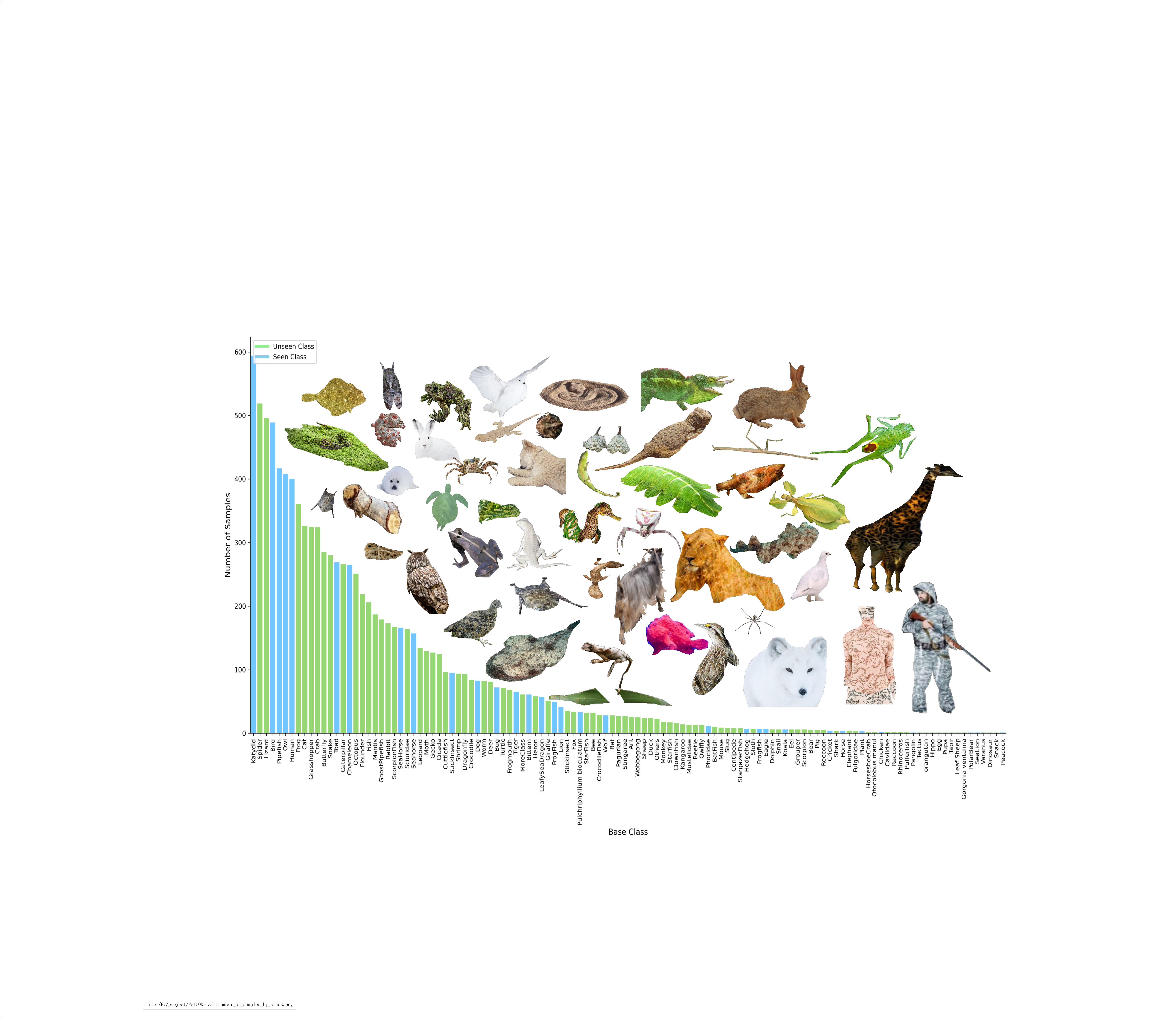}
\caption{Class distribution in the proposed \datasetname\ dataset.}
\label{fig:chart}
\vspace{-0.4cm} 
\end{figure}

\section{Proposed \datasetname\ Dataset}
\label{sec:dataset}

In order to advance the study of visual-textual approaches in COD, we propose a new dataset, {\datasetname}.
The focus of this research is to improve object detection in camouflaged scenes by integrating the class-level textual information of objects. This method extends the traditional paradigm of COD and provides a new benchmark. It should be emphasized that our innovation lies not only in the class annotation itself, but also in how to use class annotation to provide COD with new perspective and idea in visual-textual interaction without changing the existing COD model training paradigm.

\noindent\textbf{Annotation Strategy.}
\datasetname~contains 10,523 samples from 128 class, of which 44 are multi-class samples.
Each sample consists of an image (\({I}\)), ground truth (\({G}\)), and edge information (\({E}\)).
The number of images in each class is uneven (as illustrated in Fig.~\textcolor{red}{\ref{fig:chart}}), which reflects  the real-world frequency of different camouflaged objects, but this imbalance also poses challenges for the model's generalization and robustness.
In the process of labeling, we adopt strict class definition and standardization strategy to reduce ambiguity and ensure clarity and consistency of class definition, and unify the treatment of fine-grained class such as ``small fish'' and ``black cat''.
Grouping them into broader class (such as ``fish'' and ``cat'') simplifies the class space and makes it easier to identify object. 
For images containing multiple-class objects, we provide an independent segmentation mask for each class.
For example, an image containing both a "dog" and a "cat" will have separate masks for each class, ensuring accurate and high-quality annotation.

\noindent\textbf{Flexible Application.}
\datasetname~is highly flexible and extensible.
Researchers can continue to use the training and test partitioning of the original COD dataset to maintain comparability with previous studies and ensure that the introduction of new information does not affect existing experimental paradigms.
In addition, we integrate these sub-datasets to form a large camouflaged object dataset with  class information, and further divide the test set into seen classes $\mathcal{C}_{seen}$ (present in the training set) and  $\mathcal{C}_{unseen}$ (not present in the training set) to test  the generalization of COD model on  $\mathcal{C}_{seen}$ classes. The detailed criteria for this division can be found in the supplementary material. 
\datasetname\ dataset provides sufficient training data for the CGCOD task, supporting the joint semantic modeling of text and image, and the location of camouflaged object in complex scenes.
It is expected that this large-scale dataset will become a cornerstone of community research, driving the development of the COD field in open-vocabulary~\cite{OVCOS}, referring~\cite{referring}, and zero-shot learning~\cite{zero-shot_survey}.


\begin{figure*}[!t]
\centering  
\includegraphics[width=\linewidth]{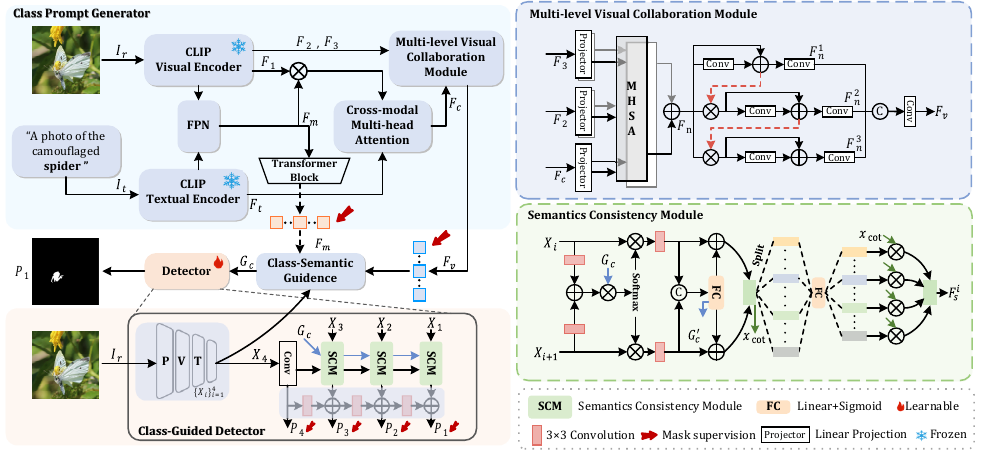}
\caption{Overview of the proposed \methodname.}
\label{fig:net}
\vspace{-13pt}
\end{figure*}

\section{Proposed Method}
\label{sec:proposed_method}

When students learn in a classroom, they do not merely rely on rote memorization but actively externalize information by recording key concepts. This focus on clearly targeted content helps them to grasp the learning points in a targeted way. Inspired by this, we introduce the \methodname\ to alleviate the limitations faced by the purely visual COD model. As illustrated in Fig.~\textcolor{red}{\ref{fig:net}},
the input of the \methodname\ is composed of tow parts, which include image and class-level textual information of the camouflaged object, termed as \(I_{r}\) and \(I_{t}\). They come from the dataset \datasetname, which are fed into the \CPGname, class-semantic guidance and \Detetorname\ to generate class-semantic prompt feature \(G_{c}\) and the  segmentation result  \(P_{1}\). 
The class-semantic guidance consists of two MHSA modules to reduce domain difference among feathers from different sources.


\subsection{Class Prompt Generator (CPG)}
\label{sec:cpg}

\begin{figure}[!t]
    \centering
    \includegraphics[width=0.8\linewidth]{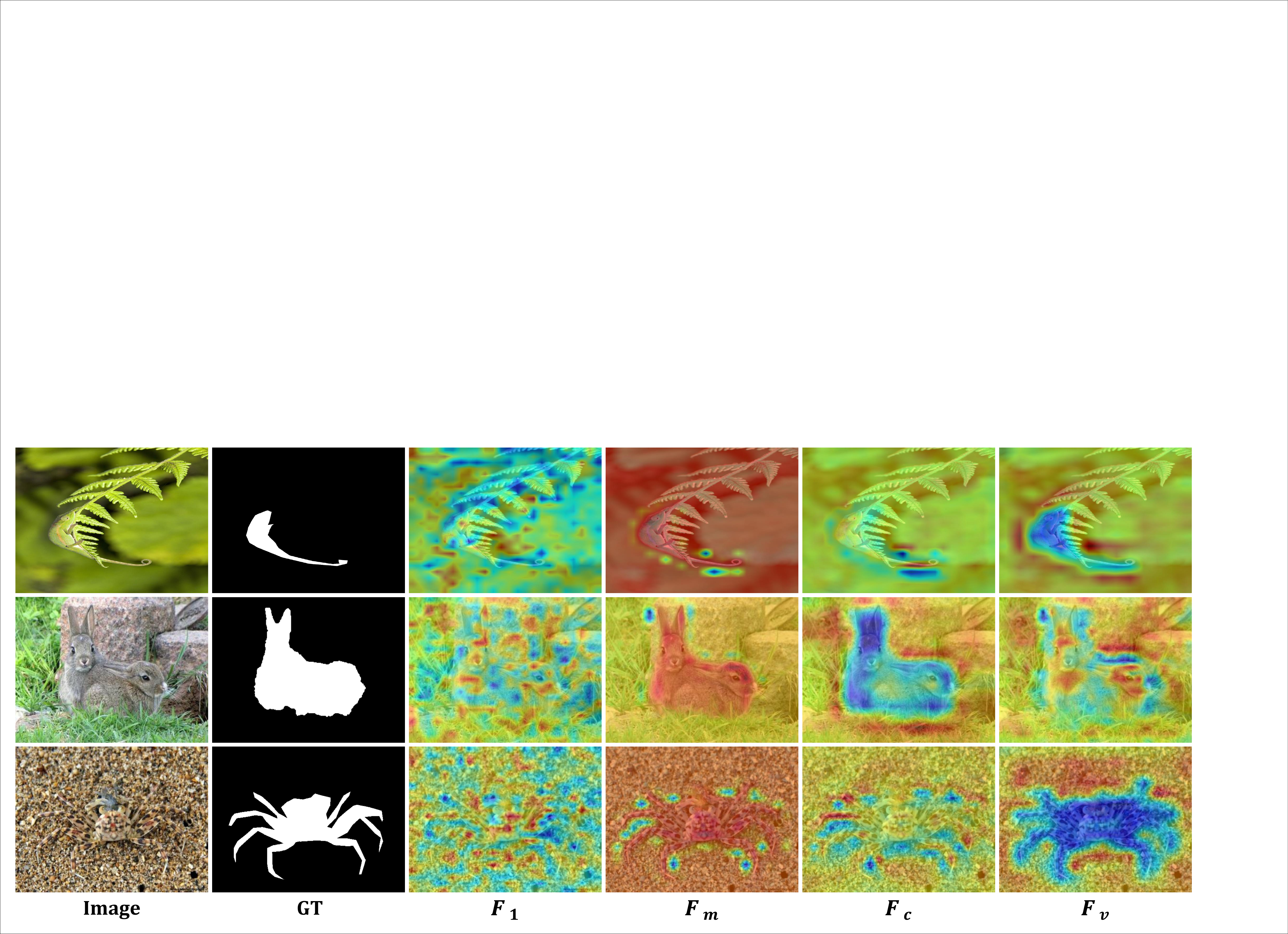}
    \caption{Visualization of the internal feature in the CPG.}
    \label{fig:CPG}
 \vspace{-0.5cm} 
\end{figure}

\noindent\textbf{Overview.}
The CPG is designed to integrate both visual and textual information to generate prompt feature for COD detector. Integrating textual information allow the model to capture deeper contextual semantics of  objects and effectively reduce interference from visually similar yet semantically irrelevant features.
This mechanism significantly suppresses both false positives and false negatives, thereby improving the precision of the COD task. 
Specifically, in this phase, we first employ a frozen CLIP textual encoder to generate textual embeddings $F_t$ from the class-level textual information. Simultaneously, we extract multi-level visual features $\{ F_i \}_{i=1}^3$ from 8, 16, and 24 layers of the frozen CLIP visual encoder. These features are then fused via a FPN module \cite{CRIS} and decode through a Transformer block to generate preliminary multi-modal features $F_m$.
To further strengthen the alignment between text and pixels and emphasize class semantic features of object, we introduce a cross-modal multi-head attention (CMA). The computation process follows the basic attention mechanism, projecting features from both modalities into different embedding spaces, and calculating the correlation between text and image features across multiple subspaces. This highlights the class semantic knowledge of object in the image, ultimately generating the feature $F_c$.  
Furthermore, considering the effectiveness of multi-level feature collaboration in image understanding, we propose a multi-level visual collaboration module (MVCM) to enrich visual features at different levels and generate the deep multimodal feature $F_v$.
Fig.~\textcolor{red}{\ref{fig:CPG}} visualizes the stages within the CPG, demonstrating progressive improvements in object localization and validating its effectiveness.

\noindent\textbf{Multi-level Visual Collaboration Module (MVCM).}
Existing research has shown that the collaboration between low and high level structural information is essential for fine-grained segmentation, reflecting principles akin to human visual processing. In order to take advantage of the coarse segmentation multi-level features of the CLIP visual encoder to generate more fine-grained visual features, we propose the MVCM to enhance the multi-layer features $F_c$, $F_2$ and $F_3$,  where $F_c$ represents low-level multimodal features, and $F_2$ and $F_3$ correspond to high-level visual features. 

To better understand and integrate information from different sources, this module consists of two stages: the alignment stage and the enhancement stage. In the {alignment} stage, we employ two multi-Head self-attention (MHSA) modules to align $F_c$ with $F_2$ and $F_c$ with $F_3$, followed by a summation operation to fuse them, generating intermediate features $F_n$. 
\begin{align}
F_{{n}} = {MHSA}(F_c, F_2) + {MHSA}(F_c, F_3)
\end{align}
In the {enhancement} stage, we apply a residual-based progressive refinement strategy on $F_{{n}}$. This strategy progressively strengthens the feature, ultimately producing the deep multimodal  fused feature representation $F_v$. This design aims to maximize the retention of fine details while extracting rich semantic information, thereby improving the model's ability to detect camouflaged object. which are formulated as:
\begin{align}
&F_n^1={Conv_3}({Conv_3}(F_n) +  F_n)\\
&F_n^2 =  {Conv_3}( {Conv_3}(F_n \otimes F_n^1) + F_n \otimes F_n^1 )\\
&F_n^3 =  {Conv_3}( {Conv_3}(F_n \otimes F_n^2) + F_n \otimes F_n^2 )\\
&F_v = {Conv_3}({Concat}(F_n^1, F_n^2, F_n^3)).
\end{align}
where $\otimes$, $Conv_3$, and $Concat$ denote Hadamard multiplication, a 3 × 3 convolutional layer, and the concatenation operation, respectively.

\subsection{Class-Guided Detector (CGD)}
\label{sec:cgnet}

\noindent\textbf{Overview.}
To improve the performance of COD, we further propose a novel detector CGD.
Which leverages class-promt information to accurately segment camouflaged objects and suppress background noise.
Given an input \( I_r \) of size \( W \times H \), We first use PvTv2-B2~\cite{pvt} as the visual backbone to extract a set of depth features \(\{X_i\}_{i=0}^{4} \).
As shown in Fig.~\textcolor{red}{\ref{fig:net}}, the semantic-rich feature \( X_4 \), the feature \( F_v \), and the initially fused multi-modal information \( F_m \) are input into the class-semantic guidance (CSG).
The module aims to explore the internal correlations between the backbone features and the other modalities from both shallow and deep layers, thus generating  prompt feature \( G_c \) that embed both class semantic and camouflaged semantic.
These features \( \{X_i\}_{i=1}^{4} \) and  \( G_c \) are then passed to the semantics consistency module to further extract fine-grained features related to the camouflaged object, ultimately leading to an accurate segmentation.
The process iteratively accumulates intermediate segmentation results to decouple the final output \( \{P_i\}_{i=1}^{4} \).

\noindent\textbf{Semantics Consistency Module (SCM).}
Traditional COD models rely on a visual backbone that extracts camouflage-semantic features containing both foreground and background clues, but this often results in false negatives and false positives.
To address these issues, we propose the SCM.
By the guidence of class-semantic prompt feature $G_c$, the SCM pays more attention to features with high correlation with guided feature, alleviating the interference of irrelevant noise information.
Camouflage mainly hides objects by manipulating channel information (e.g., colors), and in rare cases, it also uses spatial positioning to achieve the concealment effect.
So we design tow stages to decouple camouflage strategy, the first stage, spatial localization, focuses on capturing spatial information of visual features related to class-semantic prompt feature, while the second stage, channel refinement, explore channel relationship between features and adaptively selects channel information that align with class semantic to suppress noise information interference.
Specifically, we utilize two parallel branches to process multi-level features $X_i$ and $X_{i+1}$ independently, facilitating their interaction with class semantic knowledge.
Each branch incorporates weighted summation and dot product operations, ensuring a comprehensive interaction between the object class-semantic prompt feature and the multi-level features.
This interaction ensures a preliminary semantic consistency between features and class prompt feature.
Consequently, it produces $G'_{{c}}$ and enriched class semantic features $\tilde{X}_{i+1}$ and $\tilde{X}_{i}$.
The process is formulated as follows:
\begin{align}
&R_{cls} = {Softmax}\left({Conv_3}\big(X_i \ + X_{i+1}\big) \otimes G_{c}\right) \\
&\hat{X}_i = R_{cls} \otimes X_i, \quad \hat{X}_{i+1} = R_{cls} \otimes X_{i+1} \\
&G'_{c} = {FC}\left({Concat}\big(\hat{X}_{i+1},  \hat{X}_{i}\big)\right) \\
&\tilde{X}_i = G'_{c} \ + \hat{X}_i, \quad \tilde{X}_{i+1} = G'_{c} + \hat{X}_{i+1}
\end{align}
Here, \(i\) denotes the \(i\)-th operator, \( {FC} \) refers to linear projection,
while \( {Conv_3} \) signifies a \( 3 \times 3 \) convolution.

\begin{table*}[!t]
\centering
\caption{
Comparison with recent COD and RIS methods with different training settings.
``+'': optimizing the detector under our \CPGname on the proposed dataset \datasetname;
``-'': methods without publicly available results;
CGD: the proposed \Detetorname;
2D Prompt: domain-specific prompts~\protect\cite{VSCode};
COD-TAX: a coarse-grained text set generated by GPT-4V for camouflaged environments~\protect\cite{ACUMEN}.
The best three results are highlighted in \protect\textcolor{red}{\textbf{red}}, \textcolor{green}{\textbf{green}}, and \textcolor{blue}{\textbf{blue}} fonts.
}
\label{tab:metrics}
\resizebox{\linewidth}{!}{

\begin{tabular}{l|c|c|cccc|cccc|cccc|cccc}
\toprule
\multirow{2}{*}{Methods} &
\multirow{2}{*}{Prompt} &
\multirow{2}{*}{Backbone} &
\multicolumn{4}{c|}{CAMO (250 images)} &
\multicolumn{4}{c|}{COD10K (2026 images)} &
\multicolumn{4}{c|}{NC4K (4121 images)} &
\multicolumn{4}{c}{CHAMELEON (76 images)} \\
&
&
& $S_m\uparrow$ & $F^{\omega}_{\beta}\uparrow$ & $\mathcal{M}\downarrow$  & $E_{\phi}^m\uparrow$
& $S_m\uparrow$ & $F^{\omega}_{\beta}\uparrow$ & $\mathcal{M}\downarrow$  & $E_{\phi}^m\uparrow$
& $S_m\uparrow$ & $F^{\omega}_{\beta}\uparrow$ & $\mathcal{M}\downarrow$  & $E_{\phi}^m\uparrow$
& $S_m\uparrow$ & $F^{\omega}_{\beta}\uparrow$ & $\mathcal{M}\downarrow$  & $E_{\phi}^m\uparrow$ \\

\midrule
\multicolumn{19}{c}{Referring Image Segmentation} \\
\midrule
CRIS~\cite{CRIS}   & \datasetname & CLIP       & 0.305 & 0.132 & 0.457 & 0.366 & 0.258 & 0.082 & 0.583 & 0.281 & 0.278 & 0.129 & 0.533 & 0.311 & 0.251 & 0.117 & 0.581 & 0.273 \\
CGFormer~\cite{CGFormer} & \datasetname & BETR, Swin      & 0.285 & 0.177 & 0.606 & 0.285 & 0.273 & 0.086 & 0.632 & 0.262 & 0.279 & 0.147 & 0.617 & 0.276 & 0.293 & 0.133 & 0.597 & 0.286 \\
ASDA~\cite{ASDA}   & \datasetname & CLIP       & 0.790 & 0.643 & 0.091 & 0.819 & 0.807 & 0.601 & 0.048 & 0.863 & 0.847 & 0.701 & 0.059 & 0.878 & 0.832 & 0.647 & 0.063 & 0.864 \\

\midrule
\multicolumn{19}{c}{Camouflaged Object Detection} \\
\midrule
SINet~\cite{SINet}    & None & ResNet       & 0.751 & 0.606 & 0.100 & 0.771 & 0.751 & 0.551 & 0.051 & 0.806 & 0.808 & 0.723 & 0.058 & 0.887 & 0.869 & 0.740 & 0.044 & 0.899 \\
PFNet~\cite{PFNet}    & None & ResNet       & 0.782 & 0.695 & 0.085 & 0.855 & 0.800 & 0.660 & 0.040 & 0.877 & 0.829 & 0.745 & 0.053 & 0.887 & 0.882 & 0.810 & 0.033 & 0.942 \\
BGNet~\cite{BGNet}    & None & Res2Net      & 0.812 & 0.749 & 0.073 & 0.870 & 0.831 & 0.722 & 0.033 & 0.901 & 0.851 & 0.788 & 0.044 & 0.907 & 0.901 & 0.851 & 0.027 & 0.943 \\
ZoomNet~\cite{zoomnet2022}  & None & ResNet       & 0.820 & 0.752 & 0.066 & 0.877 & 0.838 & 0.729 & 0.029 & 0.888 & 0.853 & 0.784 & 0.043 & 0.896 & 0.902 & 0.845 & 0.023 & 0.952 \\
FEDER~\cite{FEDER2023}   & None & Res2Net      & 0.836 & 0.807 & 0.066 & 0.897 & 0.844 & 0.748 & 0.029 & 0.911 & 0.862 & 0.824 & 0.042 & 0.913 & 0.887 & 0.834 & 0.030 & 0.946 \\
FSPNet~\cite{FSPNet}   & None & VIT          & 0.851 & 0.802 & 0.056 & 0.905 & 0.850 & 0.755 & 0.028 & 0.912 & 0.879 & 0.816 & 0.035 & 0.914 & 0.908 & 0.868 & 0.022 & 0.943 \\

HitNet~\cite{HitNet}   & None & PVT          & 0.849 & 0.809 & 0.055 & 0.906 & 0.871 & \textcolor{green}{\textbf{0.806}} & 0.023 & 0.935 & 0.875 & 0.834 & 0.037 & 0.926 & \textcolor{green}{\textbf{0.921}} & \textcolor{green}{\textbf{0.897}} & \textcolor{green}{\textbf{0.019}} & \textcolor{green}{\textbf{0.967}} \\
PRNet~\cite{PRNet}    & None & PVT          & 0.872 & 0.831 & 0.050 & 0.922 & 0.874 & \textcolor{blue}{\textbf{0.799}} & 0.023 & \textcolor{green}{\textbf{0.937}} & 0.891 & \textcolor{blue}{\textbf{0.848}} & \textcolor{blue}{\textbf{0.031}} & 0.935 & 0.914 & 0.874 & \textcolor{blue}{\textbf{0.020}} & \textcolor{blue}{\textbf{0.971}} \\
CamoFormer~\cite{Camoformer}     & None & PVT      & 0.872 & 0.831 & 0.046 & 0.929 & 0.869 & 0.786 & 0.023 & 0.932 & \textcolor{blue}{\textbf{0.892}} & 0.847 & \textcolor{green}{\textbf{0.030}} & 0.939 & 0.910 & 0.865 & 0.022 & 0.957 \\
FSEL~\cite{FSEL}     & None & PVT      & \textcolor{blue}{\textbf{0.885}} & \textcolor{green}{\textbf{0.857}} & \textcolor{blue}{\textbf{0.040}} & \textcolor{green}{\textbf{0.942}} & \textcolor{blue}{\textbf{0.877}}  & \textcolor{blue}{\textbf{0.799}} & \textcolor{green}{\textbf{0.021}} & \textcolor{green}{\textbf{0.937}} & \textcolor{blue}{\textbf{0.892}} & \textcolor{green}{\textbf{0.852}} & \textcolor{green}{\textbf{0.030}} & \textcolor{green}{\textbf{0.941}} & 0.916 & 0.880 & 0.022 & 0.958 \\
RISNet~\cite{RISNet}   & Depth Map & PVT          & 0.870 & 0.827 & 0.050 & 0.922 & 0.873 & \textcolor{blue}{\textbf{0.799}} & 0.025 & 0.931 & 0.882 & 0.834 & 0.037 & 0.926 & 0.908 & 0.863 & 0.024 & 0.951 \\
VSCode~\cite{VSCode}  & 2D Prompt & Swin     & 0.873 & 0.820 & 0.046 & 0.925 & 0.869 & 0.780 & 0.025 & 0.931 & 0.882 & 0.841 & 0.032 & 0.935 & - & - & - & - \\
DSAM~\cite{DSAM}     & Depth Map & SAM, PVT     & 0.832 & 0.794 & 0.061 & 0.913 & 0.846 & 0.760 & 0.033 & 0.931 & 0.871 & 0.826 & 0.040 & 0.932 & - & - & - & - \\
ACUMEN~\cite{ACUMEN}   & COD-TAX & CLIP       & \textcolor{green}{\textbf{0.886}} & \textcolor{blue}{\textbf{0.850}} & \textcolor{green}{\textbf{0.039}} & \textcolor{blue}{\textbf{0.939}} & 0.852 & 0.761 & 0.026 & 0.930 & 0.874 & 0.826 & 0.036 & 0.932 & - & - & - & - \\

\midrule
\rowcolor{customcolor}
SINet+~\cite{SINet}   & \datasetname & CLIP, ResNet & 0.870 & 0.807 & 0.047 & 0.920 & 0.857 & 0.733 & 0.029 & 0.912 & 0.882 & 0.806 & 0.038 & 0.925 & 0.882 & 0.795 & 0.036 & 0.927 \\
\rowcolor{customcolor}
PFNet+~\cite{PFNet}   & \datasetname & CLIP, ResNet & 0.873 & 0.819 & 0.045 & 0.925 & 0.860 & 0.742 & 0.028 & 0.919 & 0.890 & 0.817 & 0.035 & 0.925 & 0.889 & 0.816 & 0.031 & 0.947 \\
\rowcolor{customcolor}
BGNet+~\cite{BGNet}   & \datasetname & CLIP, Res2Net & 0.879 & 0.836 & 0.042 & 0.929 & 0.875 & 0.788 & \textcolor{blue}{\textbf{0.022}} & \textcolor{blue}{\textbf{0.936}} & \textcolor{green}{\textbf{0.893}} & 0.845 & \textcolor{green}{\textbf{0.030}} & 0.939 & 0.911 & 0.858 & 0.025 & 0.951 \\
\rowcolor{customcolor}
FSPNet+~\cite{FSPNet}  & \datasetname & CLIP, VIT    & {0.879} & 0.832 & 0.041 & 0.939 & \textcolor{green}{\textbf{0.878}} & 0.768 & 0.025 & 0.933 & 0.881 & 0.830 & 0.032 & \textcolor{blue}{\textbf{0.940}} & 0.911 & 0.870 & 0.022 & 0.956 \\
\rowcolor{customcolor}
GGD (Ours) &None&PVT&0.876 & 0.838 & 0.045  & 0.932 &
0.872 & 0.793 & 0.023  & 0.935 &
0.889 & 0.847 & 0.032  & 0.935&
\textcolor{blue}{\textbf{0.917}} & \textcolor{blue}{\textbf{0.876}} & 0.021  & 0.965\\
\rowcolor{customcolor}
 \methodname~(Ours) & \datasetname & CLIP, PVT
& \textcolor{red}{\textbf{0.896}} & \textcolor{red}{\textbf{0.864}} & \textcolor{red}{\textbf{0.036}} & \textcolor{red}{\textbf{0.947}}
& \textcolor{red}{\textbf{0.890}} & \textcolor{red}{\textbf{0.824}} & \textcolor{red}{\textbf{0.018}} & \textcolor{red}{\textbf{0.948}}
& \textcolor{red}{\textbf{0.904}} & \textcolor{red}{\textbf{0.869}} & \textcolor{red}{\textbf{0.026}} & \textcolor{red}{\textbf{0.949}}
& \textcolor{red}{\textbf{0.931}} & \textcolor{red}{\textbf{0.902}} & \textcolor{red}{\textbf{0.017}} & \textcolor{red}{\textbf{0.972}}\\
\bottomrule
\end{tabular}
}
\end{table*}

To enhance the feature expression and focus on class-relevant information, we propose a channel-adaptive refinement mechanism. Initially, features $\tilde{X}_{i+1}$  and $\tilde{X}_{i}$  are concatenated along the channel dimension to form a unified feature representation \(X_{{cot}}\). This representation is then partitioned into $j$ subgroups \(\{g_1, g_2, \dots, g_j\}\) along the channel dimension, where $j=4$.  
For each subgroup \(g_j\), its concern weight \(w_j\) is calculated through the linear layer and sigmoid activation function.
These weights modulate the importance of each subgroup by element-wise multiplication, emphasizing subgroups that are consistent with the semantics of the object class.
The weighted subgroups are subsequently fused along the channel dimension to produce the refined feature representation \(F^i_{{s}}\), which highlights the critical information associated with the object class.
The process is defined as follows:
\begin{align}
&X_{{cot}} = Conv_3({Concat}(\tilde{X}_{i}, \tilde{X}_{i+1})) \\
&\{g_1, g_2, \dots, g_j\} = {Split}(X_{{cot}}) \\
&w_j = {Gate}(g_j) \\
&g'_j = Conv_3(x_{{cot}} \otimes w_j) \\
&F^i_{{s}} = Conv_3({Concat}(g'_1, g'_2, \dots, g'_j))
\end{align}
where $Gate$ refers to tow linear projection and a sigmoid.

\subsection{Loss Function}

Since our ultimate goal is to generate a binary foreground graph, a structured loss function $\mathcal{L}_{seg} = \mathcal{L}_{{bce}} + \mathcal{L}_{{iou}}$ consisting of BCE loss ($\mathcal{L}_{{bce}}$) and IoU loss ($\mathcal{L}_{{iou}}$) can be used in the optimization process \cite{wei2020f3net}.
Specifically, the loss function can be expressed as:
\begin{align}
\mathcal{L} = \sum_{i=1}^{4} \mathcal{L}_{{seg}}(P_i, {G}) + \mathcal{L}_{{seg}}(F_v, {G})+ \mathcal{L}_{{seg}}(F_m, {G})
\end{align}
where ${F_m, F_v, P_{4}, P_{3}, P_{2}, P_{1}}$ represent the predictions generated by our model, and \({G}\) refers to the ground truth.

\section{Experiments}


\subsection{Implementation Details}
\label{sec:implementation_details}

\noindent\textbf{Data Settings.}
We used  proposed \datasetname\ dataset and traditional employ the traditional train paradigm to evaluate our approach.
The training set consists of CAMO~\cite{CAMO} and COD10K~\cite{SINetV2}, with a total of 4040 samples and their class-level textual information.
The test set consists of test samples from four benchmark datasets, totaling 8615 samples and their class-level textual information. 

\noindent\textbf{Evaluation Metrics.}
In this work, we use five widely recognized and standard metrics to assess the performance of our method, including
S-measure (\(S_{\alpha}\))~\cite{Smeasure},
E-measure ($E^{m}_{\phi}$)~\cite{Emeasure},
weighted F-measure (\(F_{\beta}^{\omega}\))~\cite{wf},
average F-measure ($F^{m}_{\beta}$)
and mean absolute error (\(M\)).

\noindent\textbf{Training Settings.}
To reduce computational load and leverage CLIP’s pre-trained knowledge \cite{xu2023side}, we keep the CLIP model frozen in our experiments, using the \text{ViT-L@336}. while the parameters of the remaining network components were learnable. Following the setup in previous studies~\cite{zoomnet2022,SINetV2}, We use the Adam optimizer with as learning rate of $1 \times 10^{-4}$ to optimize the network parameters, with a batch size of 10 and a total of 200 epochs. We apply various data augmentation techniques to the training data, including random cropping, horizontal flipping, and color jittering. Specially, during both the training and inference phases, the image input size for the CLIP model was $336 \times 336$ pixels, while the input size for the \Detetorname\ was $448 \times 448$ pixels. All experiments are conducted on a server equipped with an NVIDIA RTX 3090 (with 24GB memory).

\begin{figure}[!t]
\centering  
\includegraphics[width=\linewidth]{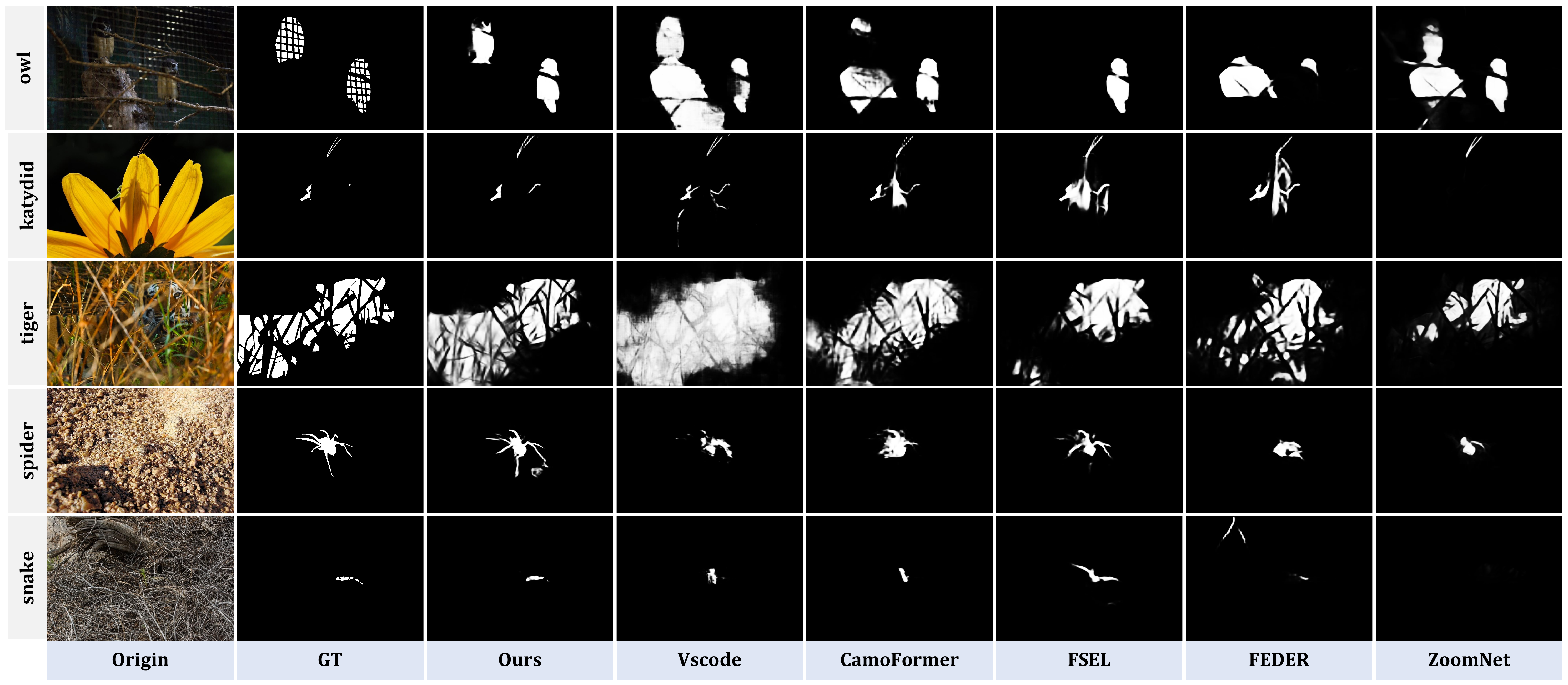}
\caption{Visual comparisons of prediction maps produced by different COD approaches.}
\vspace{-0.2cm} 
\label{fig:visual}
\end{figure}

\subsection{Comparison with State-of-the-Art Methods}

To assess our approach, we compare \methodname\ against state-of-the-art COD and RIS methods on four benchmark datasets, as summarized in Tab.~\textcolor{red}{\ref{tab:metrics}}. We also retrain certain COD and RIS methods on the \datasetname\ dataset, utilizing the CPG and CSG to gennerate prompt feature as \(X_4\)  input to enhance their performance.

\noindent\textbf{Quantitative Evaluation.}
As shown in Tab.~\textcolor{red}{\ref{tab:metrics}}, \methodname\ consistently surpasses competing methods across all datasets, showing a significant and substantial performance advantage.
And \methodname\ are on average 5.5\% better than the second best method.
Due to CRIS, CGFormer, and ASDA are not specifically designed for the COD task, their performance on the four datasets is poor.
Moreover, we also present the results of detectors optimized under the \CPGname\ (CPG) and proposed \datasetname.
In Tab.~\textcolor{red}{\ref{tab:metrics}}, CPG improve other detectors by \textbf{19.73\%} (SINet), \textbf{13.3\%} (BGNet), \textbf{12.68\%} (PFNet), \textbf{4.43\%} (FPSNet), and increase our \Detetorname\ by \textbf{6.24\%} (PVT), verifying that CPG serve as effective plug-and-play module for performance improvement. 

\begin{table}[!t] 
\centering
\caption{Comparison of methods on $\mathcal{C}_{seen}$ and $\mathcal{C}_{unseen}$.
}
\label{tab:seen}
\resizebox{0.8\linewidth}{!}{
\begin{tabular}{lcccc|cccc}
    \toprule
    \multirow{2}{*}{Methods} & \multicolumn{4}{c|}{$\mathcal{C}_{seen}$ (6109 images)} & \multicolumn{4}{c}{$\mathcal{C}_{uneen}$ (364 images)} \\
    & $S_m\uparrow$ & $F^{\omega}_{\beta}\uparrow$ & $\mathcal{M}\downarrow$ & $E_{m}\uparrow$
    & $S_m\uparrow$ & $F^{\omega}_{\beta}\uparrow$ & $\mathcal{M}\downarrow$ & $E_{m}\uparrow$ \\
    \midrule
    SINet     & 0.797 & 0.693 & 0.054 & 0.868 & 0.786 & 0.680 & 0.051 & 0.858 \\
    PFNet     & 0.820 & 0.719 & 0.050 & 0.884 & 0.801 & 0.694 & 0.051 & 0.868 \\
    BGNet     & 0.845 & 0.768 & 0.042 & 0.905 & 0.827 & 0.741 & 0.043 & 0.873 \\
    ZoomNet   & 0.849 & 0.768 & 0.039 & 0.894 & 0.826 & 0.733 & 0.043 & 0.875 \\
    FEDER     & 0.840 & 0.767 & 0.041 & 0.905 & 0.813 & 0.725 & 0.045 & 0.873 \\
    HitNet    & 0.875 & 0.827 & 0.033 & 0.929 & 0.856 & 0.797 & 0.035 & 0.916 \\
    RISNet    & 0.880 & 0.825 & 0.033 & 0.928 & 0.860 & 0.794 & 0.038 & 0.910 \\
    PRNet     & \textcolor{blue}{\textbf{0.887}}  & \textcolor{blue}{\textbf{0.834}} & \textcolor{blue}{\textbf{0.029}} & \textcolor{green}{\textbf{0.942}} & \textcolor{blue}{\textbf{0.864}} & \textcolor{blue}{\textbf{0.801}} & \textcolor{blue}{\textbf{0.035}} & \textcolor{blue}{\textbf{0.912}} \\
        FSEL      & \textcolor{green}{\textbf{0.889}} & \textcolor{green}{\textbf{0.838}} & \textcolor{green}{\textbf{0.027}} & \textcolor{blue}{\textbf{0.943}} & \textcolor{green}{\textbf{0.867}} & \textcolor{green}{\textbf{0.807}} & \textcolor{green}{\textbf{0.032}} & \textcolor{green}{\textbf{0.921}} \\
    CGD     & \textcolor{blue}{\textbf{0.887}} & 0.831 & \textcolor{blue}{\textbf{0.029}} & 0.937 & 0.861 & 0.796 & \textcolor{blue}{\textbf{0.035}} & 0.914 \\
    \rowcolor{customcolor}
    \methodname & \textcolor{red}{\textbf{0.900}} & \textcolor{red}{\textbf{0.854}} & \textcolor{red}{\textbf{0.024}} & \textcolor{red}{\textbf{0.954}} & \textcolor{red}{\textbf{0.882}} & \textcolor{red}{\textbf{0.819}} & \textcolor{red}{\textbf{0.029}} & \textcolor{red}{\textbf{0.926}} \\
    \bottomrule
\end{tabular}

}
\vspace{-0.2cm} 
\end{table}

\noindent\textbf{Qualitative Evaluation.}
Fig.~\textcolor{red}{\ref{fig:visual}} presents a visual comparison between our proposed \methodname\ method and other advanced methods.  We evaluated challenging camouflaged images featuring different object types under various extreme scenes.  These scenes include multiple objects (the first line), structural degradation (second row), occluded scenes (third row), edge similarity (fourth row), and small objects (last row). Such conditions can confound existing COD methods, leading to localization errors, blurred edges, and incomplete segmentation. As observed in Fig.~\textcolor{red}{\ref{fig:visual}}, in the first line, other advanced methods struggle to completely segment the two owls due to the high internal similarity of the samples.  In second row, the structural semantics of the katydid are disrupted, causing confusion in segmentation for other methods. In third row, heavy occlusion results in interrupted semantic information, preventing other methods from fully segmenting the object. Similarly, In forth and fifth row, complex edge details and small object sizes challenge other methods, resulting in coarse or incomplete segmentation.
In contrast to the aforementioned methods, our \methodname\ approach overcomes these challenges and demonstrates superior predictive performance across these extreme scenes.

\subsection{Ablation Study}

\begin{table}[!t]
    \centering
    \caption{Ablation comparison of proposed components.
    Baseline: PVT;
    C: semantics consistency module;
    D: CLIP;
    E: class-semantic guidance;
    F: cross-modal multi-head attention;
    G: multi-level visual collaboration module.}
    \label{tab:modules_comparison}
   \resizebox{0.8\linewidth}{!}{ 
\begin{tabular}{c|l|rrr|cccc}
    \toprule
    \multirow{2}{*}{No.}   &
    \multirow{2}{*}{Model} &
    Param.                 &
    MACs                   &
    FPS                    &
    \multicolumn{4}{c}{COD10K (2026 images)} \\
                           &            & (M)    & (G)    &       & $S_m \uparrow$ & $F^{\omega}_{\beta} \uparrow$ & $\mathcal{M} \downarrow$ & $E_{\phi}^m \uparrow$ \\
    \midrule
    I                      & Baseline   & 24.85  & 7.80   & 78.91 & 0.824                 & 0.792                         & 0.041          & 0.914               \\
    II                     & +C         & 29.08  & 17.42  & 44.40 & 0.872                 & 0.793                         & 0.023          & 0.935               \\
    III                    & +C,D       & 321.49 & 78.13  & 14.18 & 0.875                 & 0.791                         & 0.022          & 0.924               \\
    IV                     & +C,D,E     & 327.01 & 87.34  & 12.89 & 0.880                 & 0.803                         & 0.021          & 0.928               \\
    V                      & +C,D,E,F   & 342.38 & 88.94  & 12.68 & 0.889                 & 0.818                         & 0.019          & 0.937               \\
    VI                     & +C,D,E,G   & 346.78 & 90.87  & 12.48 & 0.885                 & 0.809                         & 0.020          & 0.934               \\
    VII                    & +C,D,E,F,G & 371.33 & 101.04 & 11.20 & 0.890                 & 0.824                         & 0.018          & 0.948               \\
    \bottomrule
\end{tabular}}
   \vspace{-0.1cm} 
\end{table}

\begin{figure}[!t]
    \centering
    \includegraphics[width=0.8\linewidth]{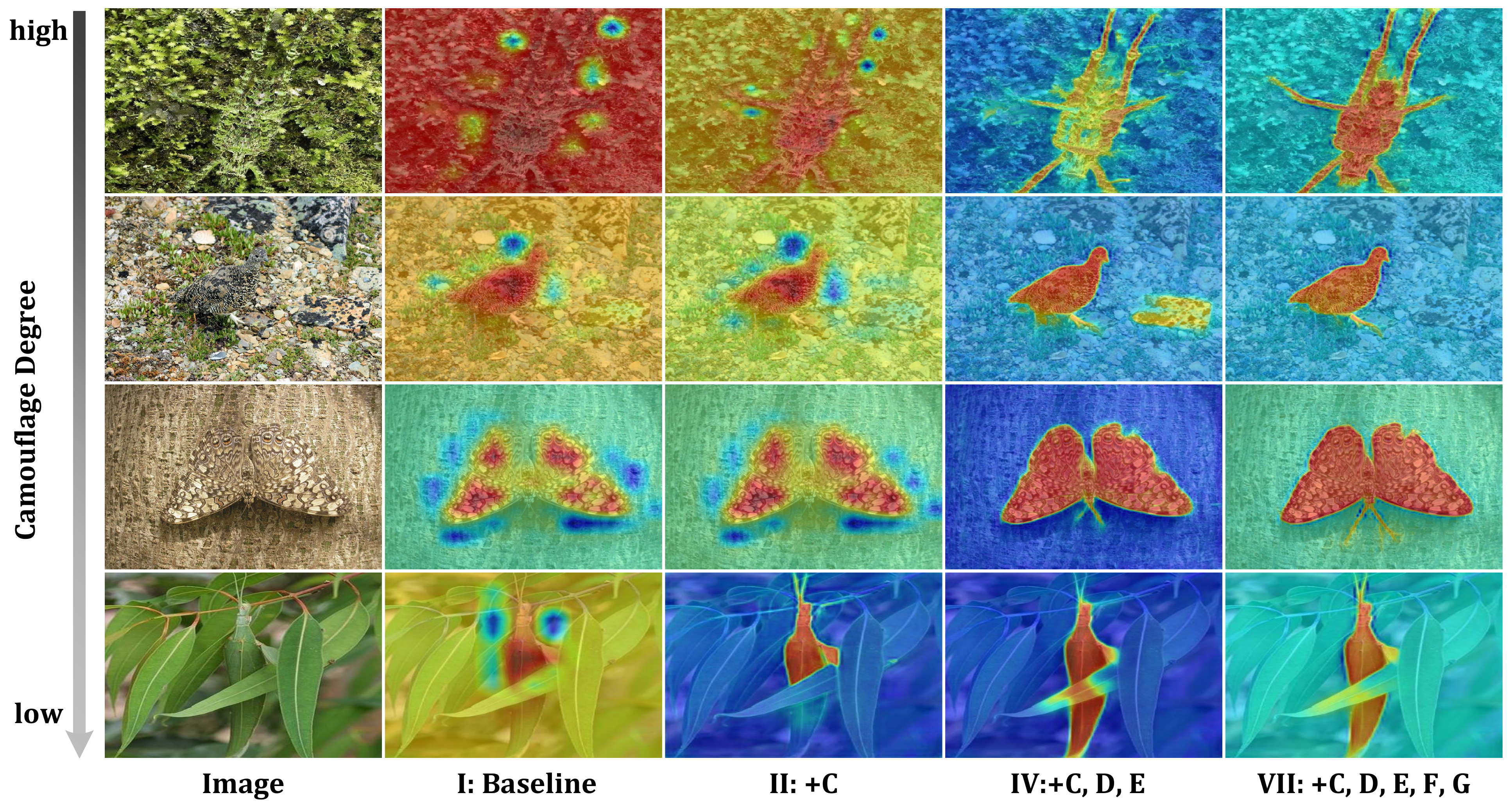}
    \caption{Visual results of the effectiveness of our components, the C, D, E, F, and G have already been explained in Tab.~\textcolor{red}{\ref{tab:modules_comparison}.}}
    \label{fig:red}
\end{figure}

\noindent\textbf{Impact of Components.}  
To analyze the effect of different components in detail, we evaluate their performance in Tab.~\textcolor{red}{\ref{tab:modules_comparison}}.
In addition, Fig.~\ref{fig:red} shows the predicted results for each component.
The groups in the table, from I to VII, are incrementally added with one module at a time, where I represents the base network.
Our analysis demonstrates that excluding any component leads to a degradation in the performance of the \methodname\ model.
II represents our proposed the \Detetorname.
Tab.~\textcolor{red}{\ref{tab:modules_comparison}} and Fig.~\textcolor{red}{\ref{fig:red}} show that CGD achieves excellent performance on COD10K.
However, it fails to completely segment objects with a high degree of camouflage.
IV performs better than II and III, which indicates the ability of class-semantic guidance to mitigate domain differences from multiple sources of visual features.
However, it still exhibits some false positives and false negatives in complex environments.
Compared to I (Baseline), VII (CGNet) improves the performance by \textbf{8.3\%}, which indicates the model progressively focuses on features that are highly relevant to the object class semantics and achieves complete object segmentation under the guidance of class-level textual information.
Particularly, incorporating the ``D'' component significantly increases the model’s parameter count, mainly due to the addition of the CLIP encoder.

\noindent\textbf{Performance on $\mathcal{C}_{seen}$ and $\mathcal{C}_{unseen}$.}
As stated in Sec.~\ref{sec:dataset}, we categorize the \datasetname~test samples into two sets $\mathcal{C}_{seen}$ and $\mathcal{C}_{unseen}$.
This division allows us to evaluate the generalization capability of different methods.
Since the distribution of $\mathcal{C}_{seen}$ and $\mathcal{C}_{unseen}$ varies significantly, we rely on results provided by the authors for a fair comparison.
As shown in Tab.~\textcolor{red}{\ref{tab:seen}}, we evaluate the performance of various COD methods on them.
Compared to other COD methods, the results demonstrate that our \methodname~achieves superior performance on both sets, particularly on $\mathcal{C}_{unseen}$.
This indicates that our framework demonstrates a certain degree of generalization ability and zero-shot capability, benefiting from the integration of class-level textual information.
Importantly, we provide more detailed analysis and comparison of visual results in the supplementary material.

\begin{table}[!t] 
\centering
\caption{Comparison of methods on hard and normal subsets.}
\label{tab:eh}
\resizebox{0.8\linewidth}{!}{\begin{tabular}{l|cccc|cccc}
    \toprule
    \multirow{2}{*}{Methods} & \multicolumn{4}{c|}{$Hard$ (6223 images)} & \multicolumn{4}{c}{$Normal$ (250 images)} \\
    & $S_m\uparrow$ & $F^{\omega}_{\beta}\uparrow$ & $\mathcal{M}\downarrow$ & $E_{m}\uparrow$
    & $S_m\uparrow$ & $F^{\omega}_{\beta}\uparrow$ & $\mathcal{M}\downarrow$ & $E_{m}\uparrow$ \\
    \midrule
    SINet     & 0.793 & 0.686 & 0.054 & 0.876 & 0.871 & 0.841 & 0.045 & 0.884 \\
    PFNet     & 0.816 & 0.712 & 0.050 & 0.892 & 0.887 & 0.856 & 0.040 & 0.897 \\
    BGNet     & 0.841 & 0.762 & 0.042 & 0.912 & 0.900 & 0.881 & 0.035 & 0.916 \\
    ZoomNet   & 0.845 & 0.762 & 0.040 & 0.910 & 0.907 & 0.884 & 0.032 & 0.918 \\
    FEDER     & 0.836 & 0.760 & 0.041 & 0.908 & 0.896 & 0.880 & 0.035 & 0.914 \\
    HitNet    & 0.872 & 0.822 & 0.033 & 0.930 & 0.912 & 0.902 & 0.028 & 0.926 \\
    RISNet    & 0.877 & 0.819 & 0.034 & 0.931 & 0.921 & 0.909 &  \textbf{\textcolor{blue}{0.027}} &  \textbf{\textcolor{blue}{0.933}} \\
    PRNet     & \textbf{\textcolor{blue}{0.884}} & \textbf{\textcolor{blue}{0.829}} & \textbf{\textcolor{blue}{0.029}} & \textbf{\textcolor{blue}{0.942}}&  \textbf{\textcolor{green}{0.924}} &  \textbf{\textcolor{blue}{0.909}} &  \textbf{\textcolor{green}{0.025}} & 0.933 \\
    FSEL      & \textbf{\textcolor{green}{0.886}} & \textbf{\textcolor{green}{0.833}} & \textbf{\textcolor{green}{0.028}} & \textbf{\textcolor{green}{0.946}} &  \textbf{\textcolor{blue}{0.923}} &  \textbf{\textcolor{green}{0.912}} &  \textbf{\textcolor{green}{0.025}} &  \textbf{\textcolor{green}{0.936}} \\
    CGD (ours) & \textbf{\textcolor{blue}{0.884}} & 0.826 & 0.030 & 0.941 & 0.921 & 0.904 & 0.028 & 0.929 \\
    \rowcolor{customcolor}
    CGNet (ours) & \textcolor{red}{\textbf{0.898}} & \textcolor{red}{\textbf{0.849}} & \textcolor{red}{\textbf{0.025}} & \textcolor{red}{\textbf{0.952}} & \textcolor{red}{\textbf{0.936}} & \textcolor{red}{\textbf{0.925}} & \textcolor{red}{\textbf{0.022}} & \textcolor{red}{\textbf{0.946}} \\
    \bottomrule
\end{tabular}
}
\end{table}

\noindent\textbf{Performance in Different Camouflaged Scenes.}
In Tab.~\textcolor{red}{\ref{tab:eh}}, we evaluate the performance of our CGNet and the existing methods in different camouflage degree scenes by splitting the \datasetname~test samples based on the detection results of CGNet.
Samples with $S_m \geq 0.9$ are categorized as $Normal$, while those with $S_m < 0.9$ are labeled as $Hard$.
The relevant analysis is detailed in the supplementary materials.

\section{Conclusion}
\label{sec:conclusion}

To overcome the limitations of the existing COD paradigm that rely solely on visual features, in this paper, we propose a novel task, termed \textbf{\taskname} (CGCOD), which employs class-level textual information to guide COD.
To facilitate this task, we meticulously construct a large-scale dataset, \textbf{\datasetname} 
and propose a novel framework CGNet for CGCOD, which include the \CPGname\ (CPG) and the \Detetorname\ .
Comprehensive experiments validate the effectiveness of the proposed CGNet, demonstrating superior performance compared to existing COD methods.
Additionally, the CPG acts as a plug-and-play component, significantly enhancing the performance of current COD detectors. 
We aim for our CGCOD research to provide valuable insights into camouflaged object perception and inspire more deep COD studies incorporating multi-modal auxiliary information.
%

{
\bibliographystyle{named}
\bibliography{ijcai25}
}

\end{document}